\documentclass[APA,Times1COL]{WileyNJDv5}
\usepackage{svg}
\usepackage{natbib}

\articletype{}

\received{Date Month Year}
\revised{Date Month Year}
\accepted{Date Month Year}
\journal{Journal}
\volume{00}
\copyyear{2023}
\startpage{1}

\newcommand{\R}{\mathbb{R}}
\newcommand{\E}{\mathbf{E}}
\newcommand{\prob}{\mathbf{P}}
\newcommand{\X}{\mathbb{X}}
\newcommand{\D}{\mathcal{D}}
\DeclareMathOperator*{\argmax}{argmax}

\newcommand*{\normaldist}{\mathcal{N}}

\newcommand*{\given}{\mid}

\newcommand*{\dd}{\,\mathrm{d}} 
\newcommand*{\mapsfrom}{\colon} 

\newcommand*{\IR}{\mathbb{R}}

\newcommand{\ninducing}{{n_\text{i}}}
\usepackage{comment}
\usepackage{multirow}
\usepackage{booktabs}
\usepackage{subcaption}
\usepackage{fancyhdr}

\begin{document}
\fancyhf{} 
\renewcommand{\headrulewidth}{0pt} 

\title{Bayesian preference elicitation for decision support in multi-objective optimization}

\author[1,2]{Felix Huber}

\author[3]{Sebastian Rojas Gonzalez}

\author[5]{Raul Astudillo}


\authormark{Huber, Rojas Gonzalez, Astudillo}
\titlemark{Bayesian preference elicitation for decision support in multi-objective optimization}

\address[1]{\orgdiv{Institute of Applied Analysis and Numerical Simulation}, \orgname{University of Stuttgart, Germany}}
\address[2]{\orgdiv{Institute for Parallel and Distributed Systems}, \orgname{University of Stuttgart, Germany}}

\address[3]{\orgdiv{Surrogate Modeling Lab}, \orgname{Ghent University - imec, Belgium}}

\address[5]{\orgdiv{Department of Computing and Mathematical Sciences}, \orgname{California Institute of Technology, USA}}

\corres{Corresponding author: Sebastian Rojas Gonzalez \email{sebastian.rojasgonzalez@ugent.be}}



\abstract[Abstract]{We present a novel approach to help decision-makers efficiently identify preferred solutions from the Pareto set of a multi-objective optimization problem. Our method uses a Bayesian model to estimate the decision-maker's utility function based on pairwise comparisons. Aided by this model, a principled elicitation strategy selects queries interactively to balance exploration and exploitation, guiding the discovery of high-utility solutions. The approach is flexible: It can be used interactively or a posteriori after estimating the Pareto front through standard multi-objective optimization techniques. Additionally, at the end of the elicitation phase, it generates a reduced menu of high-quality solutions, simplifying the decision-making process. Through experiments on test problems with up to nine objectives, our method demonstrates superior performance in finding high-utility solutions with a small number of queries. We also provide an open-source implementation of our method to support its adoption by the broader community.}


\maketitle
\thispagestyle{empty} 


\renewcommand\thefootnote{\fnsymbol{footnote}}

\section{Introduction}

Many real-world problems require optimizing multiple \emph{objectives} or \emph{criteria} simultaneously. For example, in operations management, \emph{decision-makers} (DMs) often aim to maximize expected return while minimizing risk \citep{winston2004operations}. Similarly, in aerodynamic engineering, a common goal is to maximize the lift coefficient while minimizing the drag coefficient of an airfoil design \citep{he2020}. In such cases, a single solution that optimizes all objectives rarely exists; for instance, solutions with higher expected returns typically come with increased risk. As a result, a trade-off between objectives is inevitable, and DMs need to understand this trade-off to make informed choices.

A common approach in these scenarios is to estimate the \emph{Pareto-optimal set} (also simply referred to as the \emph{Pareto set}), which consists of solutions that cannot be improved in all objectives simultaneously; the image of the Pareto set is usually termed the \emph{Pareto front}. Each solution in this set represents a different optimal trade-off between the competing objectives, providing DMs with a range of options. However, when the Pareto set is large, it becomes challenging to select a single solution for implementation \citep{cheikh2010method}. This issue is further exacerbated as the size of the Pareto set typically grows exponentially with the number of objectives, making it difficult to approximate and choose among the solutions effectively \citep{sanders2024does}.

In the literature, methods that approximate the entire Pareto set are known as \textit{a posteriori} methods and are among the most widely studied. In contrast, \emph{a priori} methods rely on DMs to specify their preferences before the optimization process begins, typically by combining the objectives into a single scalar-valued function. While this approach simplifies the selection of a final solution, it assumes that the DM can fully articulate their preferences beforehand, which is rarely the case in practice. Indeed, before the optimization process begins, DMs often have little knowledge about the trade-offs of the different criteria and the objective space in general \citep{misitano2021desdeo}.  

To address these limitations, a broad range of \textit{interactive} methods have been proposed in the literature \citep{InteractiveSurvey}. These methods aim to progressively elicit the DM's preferences during the optimization process. By presenting a series of queries, they adaptively focus on the most relevant regions of the Pareto front. While this approach can be more flexible and user-friendly, most interactive methods either make strong assumptions about the DM's underlying preferences and responses or rely on data-inefficient heuristics to select the DM queries, often leading to suboptimal or biased solutions in practice.

In this paper, we propose a novel approach to support decision-making in multi-objective optimization, addressing several major drawbacks of existing methods. Specifically, we leverage non-parametric Bayesian preference learning to model the DM's preferences, along with a principled elicitation strategy to select queries. Our approach offers three key advantages over existing methods:

\begin{enumerate} 
\item It does not rely on strong assumptions about the DM's utility function and can accommodate noisy responses. 
\item It is query-efficient, providing a mechanism to balance exploration and exploitation while effectively eliciting the DM's utility function. 
\item It offers a natural scheme for generating a menu of user-specified size, consisting of diverse high-quality solutions, at the end of the elicitation phase. \end{enumerate}

Overall,  our work provides principled and practical framework for decision support in general multi-objective problems via efficient and robust elicitation of the DM preferences. We also provide an implementation of our method to reproduce our experiments and support its use by the broader community\footnote{\url{https://github.com/qres/BPE4MOO}}.

\section{Related Work}

The literature on incorporating preference information into multi-objective optimization is extensive, with contributions spanning multiple fields such as multi-attribute utility theory, evolutionary algorithms, and Bayesian optimization (see, e.g., \citealt{wallenius2008multiple}, \citealt{InteractiveSurvey}, and \citealt{abdolshah2019multi}). These research areas have traditionally evolved independently for the most part, leading to fragmented methodologies and inconsistent terminology. Despite recent attempts to bridge these fields (e.g., \citealt{branke2016mcda}), significant challenges remain, particularly in terms of integrating preference information in a principled manner. Below, we discuss these various lines of research, focusing on connections to our work.

Originating from the operations research community, multi-attribute utility theory provides frameworks for eliciting and modeling the DM's preferences \citep{raiffa1993, Miet99, abbas2018foundations}. These methods typically involve constructing a utility function that captures the DM's preferences across multiple attributes or criteria. However, as \cite{Miet99} highlights, most traditional approaches focus on problems with discrete or finite decision spaces, limiting their applicability to continuous, high-dimensional optimization problems. In cases involving continuous decision spaces, existing methods often yield a single estimate of the DM's utility function, which can be problematic due to the substantial uncertainty that typically remains regarding the DM's preferences.

Beyond multi-attribute utility theory, the operations research community has also explored preference-based optimization techniques that are more akin to classical gradient-based methods \citep{Miet99}. Like our approach, these methods typically assume that an underlying utility function represents the DM's preferences. However, they often make strong assumptions about its form, such as a specific parametric structure or concavity (see e.g., \citealt{zionts1983interactive,mackin2011interactive}). Moreover, they also typically require the DM to respond to more challenging queries, such as those based on marginal substitution rates, and assume that the user responses are noise-free, which is typically not true in practice. Our work circumvents these limitations by employing a Bayesian approach to preference learning, which is both non-parametric and capable of efficiently handling noisy user feedback, thus making it more broadly applicable.

The incorporation of user preferences to guide the search process in optimizing complex, multi-modal landscapes has been widely studied by the evolutionary computation community \citep{branke2008consideration, deb2010interactive, branke2016using, tomczyk2019decomposition}. These algorithms use preference information to steer the evolutionary search toward the desired regions of the Pareto front. However, most of these methods rely on heuristic mechanisms to select DM queries, which often require numerous interactions with the DM. In contrast, our approach utilizes Bayesian preference learning and decision theory to more efficiently elicit the DM's preferences. For a comprehensive review of preference-based evolutionary methods, we refer to \cite{InteractiveSurvey}. Recent advances such as \cite{taylor2024accelerated} have proposed hybrid approaches to combine Bayesian learning and evolutionary optimization more effectively.

More recently, with the growing interest in artificial intelligence for decision-making, the machine learning and related communities have developed a wide array of preference learning techniques \citep{chajewska2000utilities, boutilier2002pomdp, chu2005preference, gonzalez2017preferential, benavoli2024tutorial}. However, most of this research has been conducted outside the scope of multi-objective optimization. An exception is a line of work in Bayesian optimization \citep{garnett2023bayesian}, a well-studied framework for optimizing functions with expensive evaluations. Given its focus on sample-efficiency, the Bayesian optimization community has explored schemes to incorporate user preferences in multi-objective contexts to accelerate optimization \citep{abdolshah2019multi, astudillo20, Gaudrie2020, gibson2022guiding, lin2022preference, heidari2023data}. Other efforts include applications in engineering settings, such as \cite{lepird2015bayesian}, and interactive surrogate-based design for energy systems \citep{aghaei2022surrogate}.

Among the various research directions discussed above, our work is closely aligned with a few key papers, which we describe in more detail below. \cite{tomczyk2019decomposition} also proposed a Bayesian approach to estimate the DM's utility function, but in contrast to our work, they used a parametric prior distribution based on Chebyshev scalarizations, which were then used within an evolutionary algorithm. Queries were selected uniformly at random from populations generated by the evolutionary algorithm. Our work is also related to \cite{Taylor2021}, which alternated between preference elicitation stages, where a Bayesian preference-based optimization approach was used to find DM-preferred solutions within populations generated by the evolutionary algorithm, and evolutionary stages, where the solutions from the previous stage were used as reference points to generate the next generation of solutions. In contrast, our method directly aims to identify informative queries to elicit the DM's preferences without requiring an auxiliary evolutionary algorithm. Moreover, we model the DM's utility function directly over the objective space, rather than solely in the decision space (as done in \citealt{Taylor2021}), which, as discussed in Section~\ref{sec:decision_vs_objective}, is advantageous in most cases.

Finally, our work builds on and extends prior work by the authors. \citet{astudillo20} consider multi-objective Bayesian optimization with preference learning in settings where evaluating the objective function $f$ is expensive, with a primary focus on optimizing $f$ efficiently. In contrast, our work centers on efficiently eliciting the decision-maker’s utility function $u$ through preference queries. The preference elicitation strategy we adopt, qEUBO, was originally proposed in \cite{lin2022preference} as part of a framework that alternates between learning $f$ and $u$ in a multi-objective Bayesian optimization context. Here, we broaden the integration of multi-objective optimization and preference learning by allowing the objective functions to be known or cheaply computable. We examine both interactive and a posteriori variants of our method, and explore modeling $u$ either over the decision space or the objective space. Our main contribution is a principled and practical framework for decision support in general multi-objective problems, particularly when preferences must be elicited efficiently through interaction. We also note that qEUBO was analyzed in \cite{Astudillo2023qEUBO} in a pure preference-based Bayesian optimization setting, without the multi-objective context considered here.

\section{Problem setting}
Let $f\mapsfrom \X \to \mathbb{R}^m$ denote a continuous objective function over a compact decision space $\X \subset \mathbb{R}^d$. As is standard in multi-objective optimization, for any pair of decisions $x, x' \in \X$, we say that $x$ Pareto-dominates $x'$, denoted by $x \succ_f x'$, if $f_j(x) \geq f_j(x')$ for all $j = 1, \ldots, m$, with at least one strict inequality. The set $\X_f^* := \{x \in \X : \nexists\; x' \in \X \text{ such that } x' \succ_f x\}$ is termed the Pareto set of $f$. 

We consider the problem of assisting a DM in identifying their most preferred solution within $\X_f^*$. Specifically, we assume that the DM's preferences are governed by an underlying utility function $u\mapsfrom \mathbb{R}^m \to \mathbb{R}$, and our objective is to help the DM approximately solve the problem $\argmax_{x \in \X} u(f(x))$. We note that if $u$ is monotone non-decreasing, at least one solution to this problem is guaranteed to lie in $\X_f^*$ \citep{Miet99}. Our approach, however, does not rely on this property and can also be applied when $u$ is non-monotone.\footnote{Nonetheless, in multi-objective optimization the true utility function is typically required to be monotone to ensure consistency with Pareto compliance.}

A key challenge is that $u$ cannot be observed directly. Instead, the only information available to our algorithm comes from the DM's expressed preferences, which we assume are provided through (potentially noisy) \emph{pairwise comparisons} between solutions; while we focus on pairwise comparisons in this work, our approach can easily be easily extended to best-out-of-$q$ responses (see \citealt{Astudillo2023qEUBO} for details). Formally, we assume that we are allowed to interact with the DM up to $N$ times. At each interaction, $n = 1, \ldots, N$, our algorithm selects a pair of objective vectors $y_{n,1}, y_{n,2} \in \mathbb{R}^m$ and observes the DM's response, encoded as $r_n \in \{1, 2\}$, where $r_n = 1$ if the DM prefers $y_{n,1}$ over $y_{n,2}$, and $r_n = 2$ in the opposite case. We introduce the notation $Y_n = (y_{n,1}, y_{n,2}) \in \mathbb{R}^m \times \mathbb{R}^m$ and refer to $Y_n$ as the $n$-th query. In this work, we focus on \textit{realizable} queries, i.e., queries of the form $Y_n = (f(x_{n,1}), f(x_{n,2}))$ for some $x_{n,1}, x_{n,2} \in \X$. Similarly, we denote the corresponding decision pairs as $X_n = (x_{n,1}, x_{n,2}) \in \X^2$ and, making slight abuse of notation, we write $Y_n = f(X_n)$. 

Several related formulations have been proposed in the literature. Here, we briefly outline how our approach differs from existing ones. First, we make no specific assumptions about the form of $u$. Many prior works assume that $u$ belongs to a specific parametric family \citep{deb2010interactive,tomczyk2019decomposition} or possesses certain properties, such as concavity \citep{zionts1983interactive,mackin2011interactive}. While our method is flexible enough to approximately incorporate such information if available, we do not impose restrictive assumptions on $u$.

Furthermore, unlike most existing algorithms that assume noise-free user responses \citep{zionts1983interactive,deb2010interactive}, we explicitly account for the possibility of noisy feedback. This allows our model to better handle real-world scenarios where the DM's preferences may not be consistently expressed.

Finally, while traditional approaches often focus on identifying a single optimal solution to present to the DM, we recognize that uncertainty in the DM's preferences typically remains. To address this, we propose generating a menu of high-quality solutions, taking into account the uncertainty over $u$ that remains after interacting with the DM. By offering a diverse set of potential solutions, our approach aims to mitigate the risk associated with presenting only one option, providing the DM with a richer set of choices that better reflect their uncertain preferences.

\section{Method}
Our approach consists of three main components: A probabilistic model of the DM's utility (Section~\ref{sec:model}), a preference elicitation strategy (Section~\ref{sec:elicit}), and a menu selection strategy (Section~\ref{sec:menu}). The probabilistic model allows us to infer the solutions most preferred by the DM and is updated after each interaction. The preference elicitation strategy uses an acquisition function derived from the probabilistic model to quantify the potential benefit of asking a particular query. Finally, the menu selection strategy, applied at the end of the elicitation phase, selects a small set of high-quality decisions based on our current understanding of the DM's preferences as encoded by the model.  We discuss each of these components in detail below. In addition, Section~\ref{sec:alt_design} presents alternative design choices that yield variants of our algorithm, which may be more suitable for specific scenarios. Our approach is summarized in Algorithm~\ref{alg:opt}.

\begin{algorithm}[t]
\caption{\textbf{Bayesian preference elicitation for decision support in  multiobjective optimization} \label{alg:opt}}
\begin{algorithmic}[1]
\State{Sample initial dataset $\D_0$}
\For{$n = 1,\ldots,N$}
\State{Compute posterior distribution on $u$ given $\D_{n-1}$}
\State{Compute $Y_{n} = f(X_n)$, where $X_n \in \argmax_{X\in\X^2}\mathrm{qEUBO}_{n-1}(X)$}
\State{Observe $r(Y_n)$, the DM's response for query $Y_n$}
\State{Update dataset $\D_n= \D_{n-1} \cup \{(Y_n, r(Y_n))\}$}
\EndFor\\
\Return{$\argmax_{x_1,\ldots, x_k\in \X}\E_N[\max_{i=1,\ldots, k}u(f(x_i))]$}
\end{algorithmic} 
\end{algorithm}

\subsection{Probabilistic model of the DM's utility}
\label{sec:model}

We model the DM's latent utility function using a Gaussian process (GP) for preference learning using pairwise comparisons \citep{chu2005preference}. This is done by specifying a GP prior distribution and a likelihood function that accounts for noise in the DM's responses. Recall that $Y_n = (y_{n,1}, y_{n,2}) \in \mathbb{R}^m \times \mathbb{R}^m$ represents the $n$-th query posed to the DM and $r_n \in \{1, 2\}$ denotes the corresponding response. To model the likelihood of the DM's responses, we adopt a logistic noise model defined as follows:
\begin{equation}\label{eq:likelihood}
    \prob(r_n = i \mid Y_n) = \frac{e^{u(y_{n,i})/\lambda}}{e^{u(y_{n,1})/\lambda} + e^{u(y_{n,2})/\lambda}}, \quad i = 1, 2
\end{equation}
where $\lambda > 0$ is the noise level parameter. We note that this model can recover a noise-free likelihood by letting $\lambda \to 0$.

Let $\D_0$ denote a potentially empty initial dataset and $\D_{n-1} = \D_0\cup\{(X_l, r(X_l))\}_{l=1}^{n-1}$ denote the data collected before the $n$-th interaction with the DM. The posterior distribution over $u$ given $\D_{n-1}$ can be updated using Bayes' rule. However, due to the combination of our choice of GP prior and logistic likelihood, the posterior is not analytically tractable. Instead, it can be approximated using methods such as the Laplace approximation \citep{cheikh2010method} or variational inference \citep{Nguyen2021TopK}. In our experiments, we employ the latter approach due to its computational efficiency and scalability; see Appendix \ref{app1.1a} for a description of the approach. 

\subsection{Preference elicitation strategy}
\label{sec:elicit}
We select the queries to be presented to the DM by maximizing an acquisition function constructed from the current posterior distribution. In our experiments, we use the qEUBO acquisition function proposed by \cite{lin2022preference} and \cite{Astudillo2023qEUBO} due to its tractability, strong empirical performance, and theoretical guarantees. However, other acquisition functions could be employed as well. For completeness, we provide a brief description of the qEUBO acquisition function below.

The qEUBO acquisition function is defined as
\begin{equation} 
\mathrm{qEUBO}_{n-1}(X) = \mathbf{E}_{n-1}\biggl[\max_{i=1,2}u(f(x_i)) \biggr], 
\end{equation}
where the subscript $n-1$ indicates that the expectation is taken with respect to the posterior distribution of $u$ after $n-1$ interactions with the DM. At each interaction, we select the query to be presented by maximizing this acquisition function. Specifically, during the $n$-th interaction, we set $Y_n  = f(X_n)$, with
\begin{equation}
  X_n \in \argmax_{X \in \X^2} \mathrm{qEUBO}_{n-1}(X).
\end{equation}

We note that maximizing the acquisition function requires potentially many evaluations of $f$. In many scenarios, this is computationally feasible. 
When $f$ is expensive to evaluate, a potential alternative is to use a separate probabilistic model over $f$, as suggested by \cite{lin2022preference}. However, our focus is on problems where evaluating $f$ enough times to maximize the acquisition function is feasible.

In Section~\ref{sec:decision_vs_objective} we discuss an alternative modeling approach, where the DM's responses are used to infer the mapping $x \mapsto u(f(x))$ directly, instead of $u$. This approach eliminates the need to evaluate $f$ during the acquisition function maximization step but comes with a trade-off: It requires modeling a function over a potentially higher-dimensional decision space, since the number of decision variables $d$ is typically larger than the number of objectives $m$.

\subsection{Menu selection strategy}
\label{sec:menu}
An important advantage of our probabilistic approach over most existing interactive methods is that it provides a natural scheme for generating a menu of high-quality items to present to the DM after the elicitation phase is complete. This menu is designed to maximize the expected utility that the DM would receive when selecting the best item according to their underlying utility function, given the remaining uncertainty after the elicitation phase. Specifically, our approach selects the menu items by maximizing the expected utility of the best choice, defined as
\begin{equation}
\label{eq:menu}
\max_{x_1, \ldots, x_k \in \X} \mathbf{E}_{n}\biggl[\max_{i=1, \ldots, k} u(f(x_i))\biggr],
\end{equation}
where $k$ is the desired menu size.

It is noteworthy that, for $k=2$, the above expression coincides with the qEUBO acquisition function. This dual interpretation of qEUBO, both as an acquisition function during the elicitation phase and as a strategy for selecting menus afterward, has been previously observed in the literature \citep{viappiani2010optimal}. Although this is not critical to our setting, it intuitively implies that the queries presented by our algorithm are composed of high-quality points, similar to those we would select if we were to finalize the elicitation phase at that particular interaction.

\subsection{Alternative design choices}
\label{sec:alt_design}
Having outlined the main version of our algorithm, we now explore several alternative design choices that lead to different variants. These alternatives may be advantageous in specific scenarios and can enhance the algorithm's applicability.

\subsubsection{A priori and a posteriori versions of our algorithm}
Our algorithm, as described above, can be naturally categorized as an interactive multi-objective optimization method. It is not an a priori method, as it does not require the user to specify a single utility function representing their preferences before the optimization process begins. Similarly, it is not an a posteriori method, as it does not rely on building an estimate of the Pareto front of $f$ prior to the elicitation phase. Instead, the algorithm directly aims to find the point that maximizes the DM's utility $u(f(x))$ by interactively querying the DM. Below, we discuss how our approach can be adapted to function as both an a priori and a posteriori method in specific scenarios.

\paragraph*{A priori version with an informative prior} In certain cases, it is possible to construct an informative Bayesian prior distribution over the DM's utility function using historical data or transfer learning. When such a prior is available, our approach can be adapted to operate as an a priori method. Specifically, we can generate a menu for the DM by following the procedure described in Section~\ref{sec:menu}, computing the expectation with respect to the prior distribution without requiring any further interaction with the DM. This allows the DM to make an informed decision based on the prior knowledge alone, bypassing the need for an interactive elicitation phase.

\paragraph*{A posteriori version with a precomputed estimate of the Pareto set}
In general, accurately computing the entire Pareto set can be challenging. However, there may be scenarios where a reasonable approximation can be obtained using traditional multi-objective optimization techniques, such as evolutionary algorithms. In these situations, our approach can be used as an a posteriori method to help the DM identify their most preferred solution among points in the approximate Pareto set. Let $\widehat{\X}_f^*$ denote this approximation. This can be achieved by simply replacing $\X$ with $\widehat{\X}_f^*$ in Algorithm \ref{alg:opt} (i.e., maximizing the acquisition function and selecting the menu using only points in $\widehat{\X}_f^*$).

\subsubsection{Modeling preferences over decision vectors vs. objective vectors}
\label{sec:decision_vs_objective}

Since the DM typically expresses preferences over objective vectors $ f(x) $, it is natural to model utility as a function over the objective space, i.e., $ y \mapsto u(y) $ for $ y \in \mathbb{R}^m $. Alternatively, one can model preferences directly over the decision space by considering the composite mapping $ x \mapsto u(f(x)) $ with $ x \in \X $.

In general, modeling in objective space offers several advantages: (i) the number of decision variables often exceeds the number of objectives, making the lower-dimensional objective space more tractable; (ii) it allows for the incorporation of structural prior knowledge, such as monotonicity of the utility function (see Section~\ref{sec:monot}); and (iii) it decouples preference modeling from the complexities of the objective function, avoiding the need for the utility model to capture nonlinearities or irregularities introduced by the mapping from decision to objective space.

Nonetheless, modeling in decision space can be beneficial in specific scenarios: For instance, when the number of decision variables is smaller than the number of objectives, or when evaluating $f$ is computationally expensive and should be avoided during acquisition function optimization. This formulation also permits the use of gradient-based optimization, as a GP defined over decision space is directly differentiable. In contrast, modeling over the objective space typically requires zeroth-order optimization methods, as gradients of $f$ are often unavailable or expensive to compute.

In our experiments in Section~\ref{sec:experiments}, we compare both modeling strategies to offer practical guidance for selecting the most suitable approach depending on the problem setting.

\subsubsection{Monotonocity of the utility function}
\label{sec:monot}
In most settings, the decision-maker’s utility function $u(y)$ is known to be monotone increasing, reflecting a preference for improving all objectives. This structural assumption offers valuable prior knowledge that, if properly incorporated, can enhance both modeling accuracy and query efficiency. However, GP priors almost surely generate non-monotone sample paths, making it challenging to enforce monotonicity exactly within the model.

To address this, we adopt a simple yet effective data augmentation strategy to approximately impose monotonicity when modeling utility over the objective space. Specifically, we augment the training set with \emph{virtual} queries of the form $(Y, r(Y))$, where $Y = (y_1, y_2)$ and $y_1$ dominates $y_2$, with $r(Y) = 1$ (i.e., $y_1$ is preferred), as described in Algorithm~\ref{alg:monot}. These synthetic comparisons encourage the model to behave monotonically in the vicinity of the augmented pairs, guiding query selection toward regions closer to the Pareto front.

Our experiments show that this simple strategy can yield meaningful performance gains. Nonetheless, other techniques for enforcing monotonicity, such as constrained GPs or basis-function approximations \citep{Riihimaeki2010Gaussian, DaVeiga2012Gaussian, Maatouk2017Gaussian, LopezLopera2018FiniteDimensional}, remain a promising avenue for future work.

\begin{algorithm}[t]
\caption{Generate points to induce monotonicity}
\begin{algorithmic}[1]
\Require Dataset $\mathcal{Y}_n = \{ y_{l,i} : l = 1, \ldots, n, \  i=1,2\}$, number of objectives $m$, expansion parameter $\delta > 0$
\State Compute bounding box of current dataset $\mathcal{D}_n$:
    \State $y_{\text{min}, j} = \min_{y \in \mathcal{Y}_n} y_j$ \quad for $j = 1, \ldots, m$
    \State $y_{\text{max}, j} = \max_{y \in \mathcal{Y}_n} y_j$ \quad for $j = 1, \ldots, m$
\State Normalize objective space to $[0, 1]^m$ using $y_{\text{min}}$ and $y_{\text{max}}$
\State Draw two points $y_1, y_2$ uniformly from an expanded box \mbox{$[-\delta, 1+\delta]^m$}, such that $y_1$ dominates $y_2$
\State \Return $(y_{\text{min}} + (y_{\text{max}} - y_{\text{min}})y_1, y_{\text{min}} + (y_{\text{max}} - y_{\text{min}})y_2)$
\end{algorithmic}
\label{alg:monot}
\end{algorithm}

\section{Experiments}\label{sec:experiments}
In this section, we assess the ability of our approach to identify high-utility decisions based on a DM's unknown utility function. To this end, we use well-known synthetic test functions from the multi-objective optimization literature, as well as a practical application example, each paired with a distinct utility function (described in Section~\ref{sec:test_funcs}). Our primary experiments, detailed in Section~\ref{sec:main_results}, evaluate the performance of our approach across various decision space dimensions and number of objectives (see Table~\ref{tab:test_problems}). These experiments also explore two key design choices: Modeling the DM's preferences over decision space versus objective space, and using our approach interactively versus a posteriori. By considering each possible combination of these two design choices, we obtain four versions of our algorithm. In our plots, we label these design choices as follows:
\begin{itemize}
    \item \emph{int:} Refers to the interactive version of our algorithm.
    \item \emph{post:} Refers to the a posteriori version of our algorithm.
    \item \emph{dec:} Refers to the version of our algorithm that models the utility function over the decision space.
    \item \emph{obj:} Refers to the version of our algorithm that models the utility function over the objective space. 
\end{itemize}
Thus, for example, the interactive version of our algorithm that models the utility function over objective space is labeled as \textit{int-obj}. Additionally, we perform secondary experiments that examine the effects of approximately enforcing monotonicity in the probabilistic model (Section~\ref{sec:monot_results}), noise in the DM's responses (Section~\ref{sec:noise_results}), and varying menu sizes (Section~\ref{sec:menu_results}).

As discussed earlier, there is a plethora of methods for multi-objective optimization with preferences; comparing our approach against these methods is challenging for several reasons. For example, (1) there are many different ways of collecting and processing preference information (e.g., the frequency of interaction or type of information), (2)  the performance evaluation of these methods is often linked to the modeling assumptions and thus method-dependent \citep{afsar2021assessing}, (3) the so-called \emph{most preferred} solution can be selected in multiple ways, and (4) not all methods are scalable in both decision and objective spaces. As a comparison with an existing state-of-the-art method, we note that one of the secondary versions of our algorithm aligns closely with the method proposed by \cite{Taylor2021}. This version of our algorithm models the DM's utility over decision space and conducts the elicitation phase a posteriori over an approximation of the Pareto front (labeled as \textit{post-dec}). This can be interpreted as the algorithm in \cite{Taylor2021}, where instead of using the Expected Improvement, as proposed in their paper, all queries are collected using the qEUBO acquisition function, which has been shown to outperform Expected Improvement \citep{Astudillo2023qEUBO}. 

\subsection{Test functions}
\label{sec:test_funcs}

We evaluate the proposed framework on a suite of well-established multi-objective optimization benchmarks that present diverse characteristics and challenges \citep{huband2006review}. Specifically, we use DTLZ7 (5,3), DTLZ2 (9,6), WFG3 (14,9), and Car Cab Design (7,9), where the values in parentheses denote the dimensionality of the decision and objective spaces, respectively. This selection allows us to assess the framework’s performance across a range of problem scales and structural complexities. Formal definitions of the synthetic test functions and the Car Cab Design problem can be found in \citet{huband2006review} and \citet{lin2022preference}, respectively.

Each benchmark is paired with a distinct utility function to model the DM's preferences (see Table~\ref{tab:test_problems}). These utility functions are chosen to be monotonic over the objective space, ensuring that the global optimum lies on the Pareto set. Moreover, they represent commonly used classes of utility models across various domains. For example, DTLZ7 is coupled with a linear utility function, a widely adopted and interpretable model \citep{raiffa1993,Miet99,abbas2018foundations}. DTLZ2 uses a 3-norm utility function relative to an ideal point, inspired by \citet{tomczyk2019decomposition}. WFG3 is paired with an exponential utility function, another popular choice in the literature \citep{astudillo20,abbas2018foundations}. For the Car Cab Design problem, we adopt a piecewise linear utility function following the setup in \citet{lin2022preference}.

\begin{table}[t]
    \centering
    \begin{tabular}{@{} p{0.2\textwidth} p{0.215\textwidth} @{}}
        \toprule
        \textbf{Objective function \boldmath$(d,m)$} & \textbf{Utility function}
        \\ \midrule
        \mbox{\textbf{DTLZ7 (5,3)}}\hspace{\textwidth}
        disconnected, mixed Pareto front &
        \(\begin{aligned}[t]
        u(y) = \sum_{i=1}^m y_i
        \end{aligned}\)
        \\ \midrule
        \mbox{\textbf{DTLZ2 (9,6)}}\hspace{\textwidth}
        continuous and concave Pareto front & 
        \(\begin{aligned}[t]
            & u(y) = -\biggl( \sum_{i=1}^m (z_i - y_i)^3 \biggr)^{\frac{1}{3}}
            \\
            &\text{with } z_i = \begin{cases}
                0.2 & \text{if } i \text{ is even}, \\
                0 & \text{otherwise}
            \end{cases}
        \end{aligned}\)
        \\ \midrule
        \mbox{\textbf{WFG3 (14,9)}}\hspace{\textwidth}
        linear and degenerate Pareto front &
        \(\begin{aligned}[t]
        &u(y) = -\theta^{-1} \log \biggl( \sum_{i=1}^m \exp(-\theta y_i) \biggr)
        \\
        &\text{for } \theta = 4
        \end{aligned}\)
        \\ \midrule
        \mbox{\textbf{Car Cab Design (7,9)}}\hspace{\textwidth}
        non-convex Pareto front &
        \(\begin{aligned}[t]
        &u(y) = \sum_{i=1}^m h_i(y_i)
        \\
        &\text{with piecewise linear functions } h_i
        \end{aligned}\)
        \\ \bottomrule

    \end{tabular}
    \caption{Objective functions and their corresponding utility functions used in our empirical evaluation.}
    \label{tab:test_problems}
\end{table}

\subsection{Performance metric} \label{sec:perf_metric}

We evaluate performance using \emph{regret}, defined as the difference between the true optimal utility and the utility of the best point in a menu of size $k$ selected from a set $\mathcal{X}$:
\begin{equation*}
    \text{regret}(\mathcal{X}, k) = u(x^*) - \max_{i=1,\ldots,k} u(\widehat{x}_i^*),
\end{equation*}
where $x^* = \argmax_{x \in \X} u(x)$ is the true optimal point, and $\widehat{x}_1^*, \ldots, \widehat{x}_k^*$ are the points in $\mathcal{X}$ that maximize the expected utility of the best item in the menu:
\begin{equation*}
    (\widehat{x}_1^*, \ldots, \widehat{x}_k^*) \in \arg\max_{x_1, \ldots, x_k \in \mathcal{X}} \E_n\left[\max_{i=1,\ldots,k} u(f(x_i))\right].
\end{equation*}

In our main experiments (Section~\ref{sec:main_results}), we set $\mathcal{X} = \{x_{l,i} : l = 1, \ldots, n,\ i = 1, 2\}$, i.e., the set of all queried points up to step $n$. This ensures a fair comparison between the interactive and a posteriori versions of our algorithm, since the latter only queries points from the approximated Pareto front $\widehat{\X}_f^*$ and the model is generally unreliable outside that region. In all other experiments, we use the full decision space and set $\mathcal{X} = \X$.

In experiments that do not focus on menu selection (i.e., all except those in Section~\ref{sec:menu_results}), we set $k = 1$ to evaluate the regret of a single recommended point:
\begin{equation*}
    \text{regret}(\mathcal{X}, 1) = u(x^*) - u(\widehat{x}_1^*),
\end{equation*}
where $\widehat{x}_1^* = \arg\max_{x \in \mathcal{X}} \E_n[u(f(x))]$ is the point with the highest posterior mean utility.

\subsection{Experimental setup}

All experiments begin with an initial set $\mathcal{D}_0$ of $2(d+1)$ query pairs obtained from points drawn uniformly at random over the decision space. The DM's preferences are then elicited for each pair, forming the initial dataset $\D_0$ (line 1 in Algorithm~\ref{alg:opt}). Each algorithm is allocated a budget of 50 additional DM interactions beyond the initial set.

Our GP model of the utility function uses a Matérn covariance function with parameter $\nu = 5/2$ and a zero mean function. The covariance function's length scales, the likelihood's noise level, and the variational distribution's parameters are estimated by maximizing a variational lower bound of the log-marginal likelihood, following the approach of \cite{Hensman2015Scalable}.

In the a posteriori versions of our algorithm, we utilize standard evolutionary algorithms. Specifically, we use NSGA-II \citep{deb2002fast} for scenarios with three objectives and NSGA-III \citep{deb2013evolutionary} for scenarios with more objectives. For the cases with three objectives, we use a population of 200 instances and optimize over 500 generations. For the cases with more than three objectives, we use a population of 1500 instances and optimize over 2500 generations.

\subsection{Results} 

In our main results in Section~\ref{sec:main_results}, the simulated DM responses are generated based on the true underlying utility function without any added noise, although this information is not known to our model. In addition, our main results do not use monotonicity. In Section \ref{sec:monot_results} we discuss how enforcing monotonicity may enhance the efficiency and robustness of our method, and in Section~\ref{sec:noise_results} we show that our method remains robust even under moderate noise levels. Finally, our main results use a menu size of 1; in Section~\ref{sec:menu_results}, we explore the impact of increasing the menu size.

\subsubsection{Main results} \label{sec:main_results}

\begin{figure}
    \centering
    \captionsetup[sub]{margin={-25pt,0pt}} 
    \begin{subfigure}{0.32\textwidth}
    \caption{DTLZ7 (5,3)}
    \includegraphics[width=\textwidth]{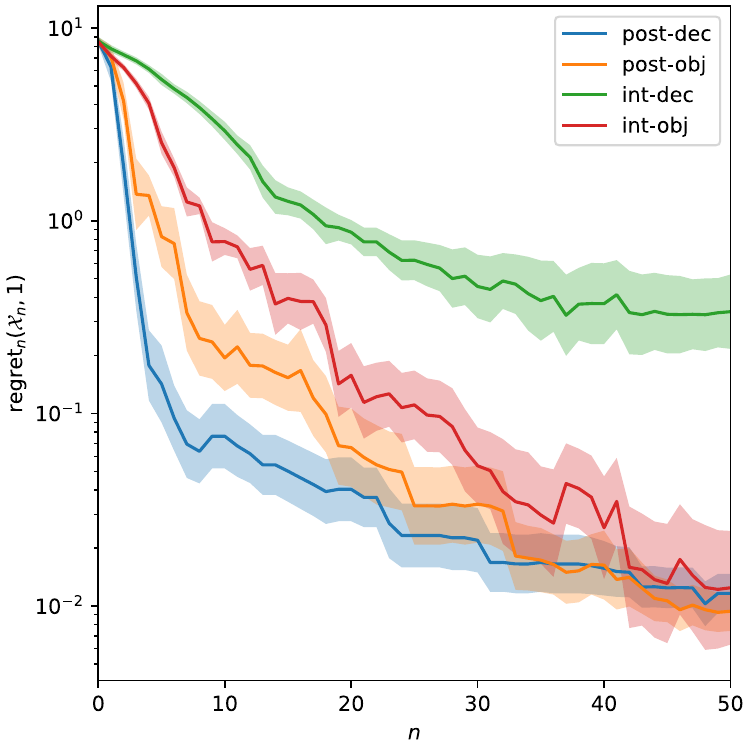}%
    \end{subfigure}
    \hspace{2.5ex}
    \begin{subfigure}{0.32\textwidth}
    \caption{DTLZ2 (9,6)}
    \includegraphics[width=\textwidth]{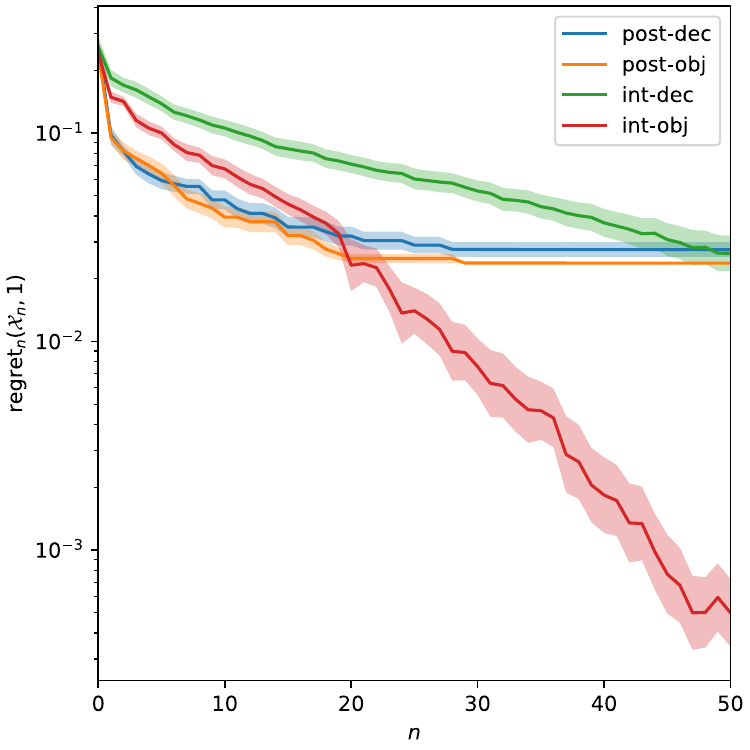}%
    \end{subfigure}
    \\[1.5ex]%
    \begin{subfigure}{0.32\textwidth}
    \caption{WFG3 (14,9)}
    \includegraphics[width=\textwidth]{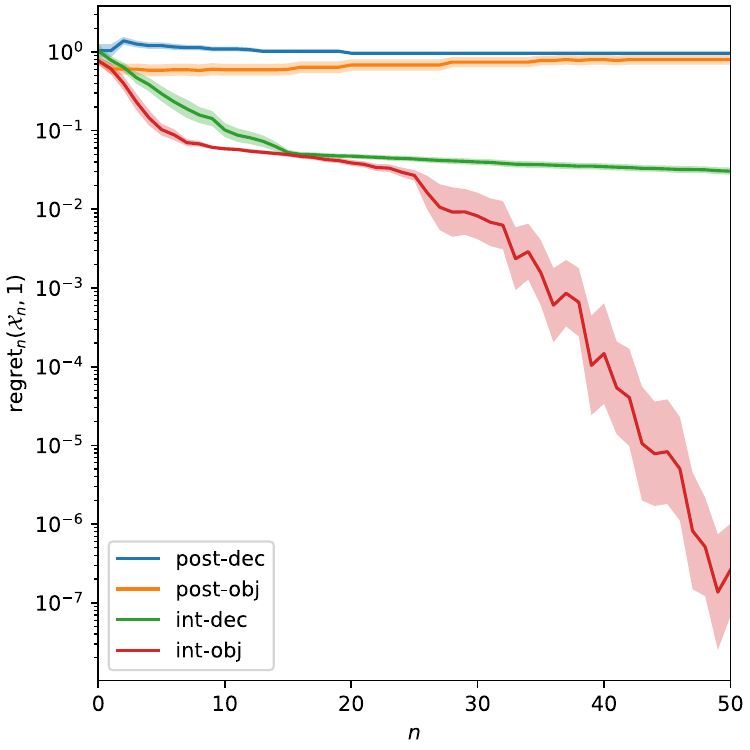}%
    \end{subfigure}
    \hspace{2.5ex}
    \begin{subfigure}{0.32\textwidth}
    \caption{Car Cab Design (7,9)}
    \includegraphics[width=\textwidth]{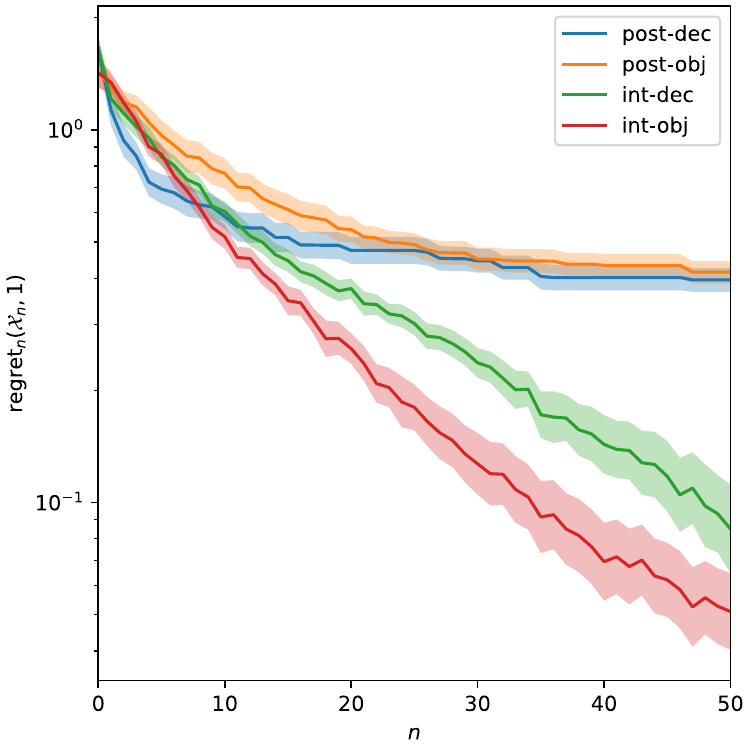}%
    \end{subfigure}
    \caption{Mean regret and standard error for DTLZ7 with 5 decision variables and 3 objectives, DTLZ2 with 9 decision variables and 6 objectives, WFG3 with 14 decision variables and 9 objectives, and Car Cab Design with 7 decision variables and 9 objectives.}
    \label{fig:scaling}
\end{figure}

 A comparison of the performance of the different algorithms is shown in Figure~\ref{fig:scaling}. Overall, the interactive algorithm that models the DM's utility over the objective space (\textit{int-obj}) achieves the best performance, particularly on problems with a larger number of objectives. We highlight below the main trends observed across the two key design choices.

First, interactive algorithms consistently outperform their a posteriori counterparts, with the exception of the DTLZ7 (5,3) problem. In this case, the small number of objectives allows a posteriori methods to begin with a high-quality approximation of the Pareto set, giving them an early advantage. However, as the number of objectives increases, accurately approximating the Pareto front becomes increasingly difficult. This leads to different performance dynamics depending on the number of DM interactions, as illustrated by the DTLZ2 (9,6) problem. A posteriori methods tend to converge quickly at first, benefiting from their initial approximation, but often plateau as this approximation lacks resolution. In contrast, interactive algorithms continue to improve by adaptively focusing on high-utility regions, enabling fine-grained refinement near the Pareto front. This trend is especially pronounced in the Car Cab Design (7,9) problem, where a posteriori methods show early gains but stagnate, while interactive methods steadily progress. The performance gap widens further in higher-dimensional settings such as WFG3 (14,9), suggesting that a posteriori approaches may be preferable only when a reliable Pareto front approximation is available.

Second, algorithms that model utility over the objective space generally outperform those that model it over the decision space. This is likely due to the lower dimensionality of the objective space, which simplifies utility modeling and improves data efficiency. An exception arises in the Car Cab Design (7,9) problem, where the objective space is higher-dimensional than the decision space; however, the performance gap remains modest. This pattern holds across both interactive and a posteriori settings, reinforcing the benefits of modeling in objective space.

\begin{figure*}
    \centering
    \includegraphics[width=1.0\linewidth]{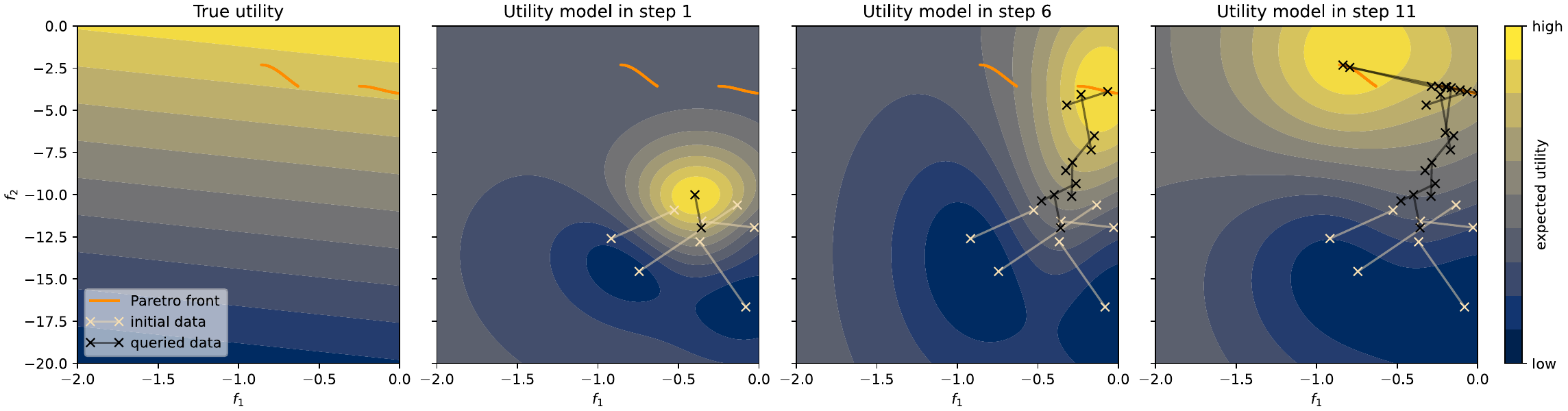}%
    \caption{Convergence of our algorithm in objective space for DTLZ7 with $5$ decision variables and $2$ objectives. The leftmost panel shows the Pareto front (orange) together with contour lines of the true utility function. The subsequent panels depict the $4$ initial queries (beige crosses), the algorithm’s selected queries (black crosses), and the model’s predicted expected utility after $1$, $6$, and $11$ user interactions. }
    \label{fig:illustration}
\end{figure*}

To illustrate the learning dynamics, Figure~\ref{fig:illustration} shows a representative run of the algorithm on the bi-objective DTLZ7 (5,2) problem. The leftmost panel shows the true Pareto front (orange) together with the contour lines of the utility function. The subsequent panels illustrate how the algorithm incrementally explores the objective space and refines its utility model. Starting from a few initial queries (beige crosses), the algorithm adaptively selects additional queries (black crosses), enabling it to rapidly identify high-utility trade-offs and converge toward the Pareto front. The learned model captures key qualitative trends of the true utility, such as higher values in regions where both $f_1$ and $f_2$ are large. However, because the model is not perfectly monotonic, these trends are only approximated. Moreover, the absolute scale of the predicted utility does not necessarily coincide with that of the true utility, since the model relies solely on pairwise comparisons rather than direct observations.

\subsubsection{Monotonocity}
\label{sec:monot_results}

\begin{figure}
    \centering
    \captionsetup[sub]{margin={0pt,-25pt}} 
    \begin{subfigure}{0.32\textwidth}
    \caption{DTLZ2 (9,6)}
    \includegraphics[width=\textwidth]{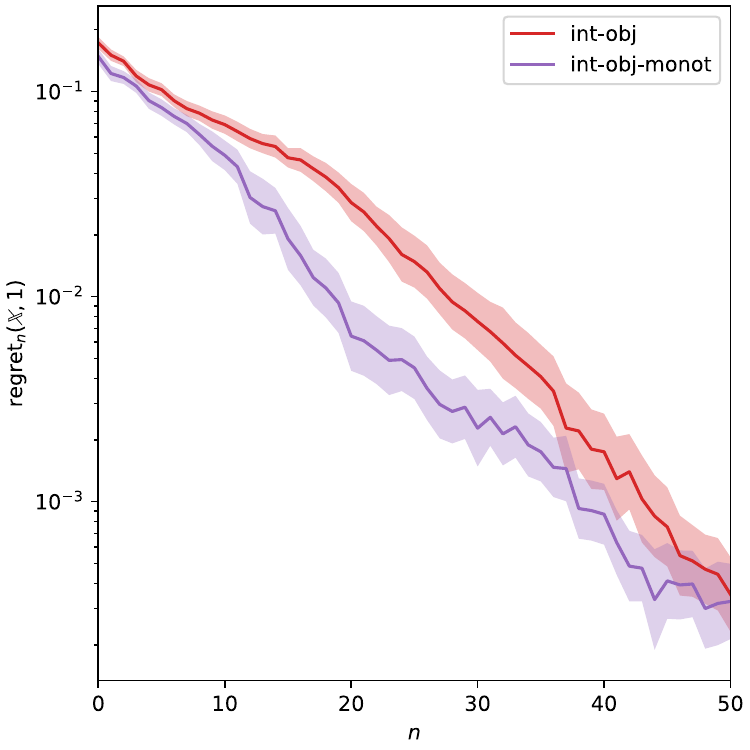}%
    \end{subfigure}
    \hfill{}
    \begin{subfigure}{0.32\textwidth}
    \caption{WFG3 (14,9)}
    \includegraphics[width=\textwidth]{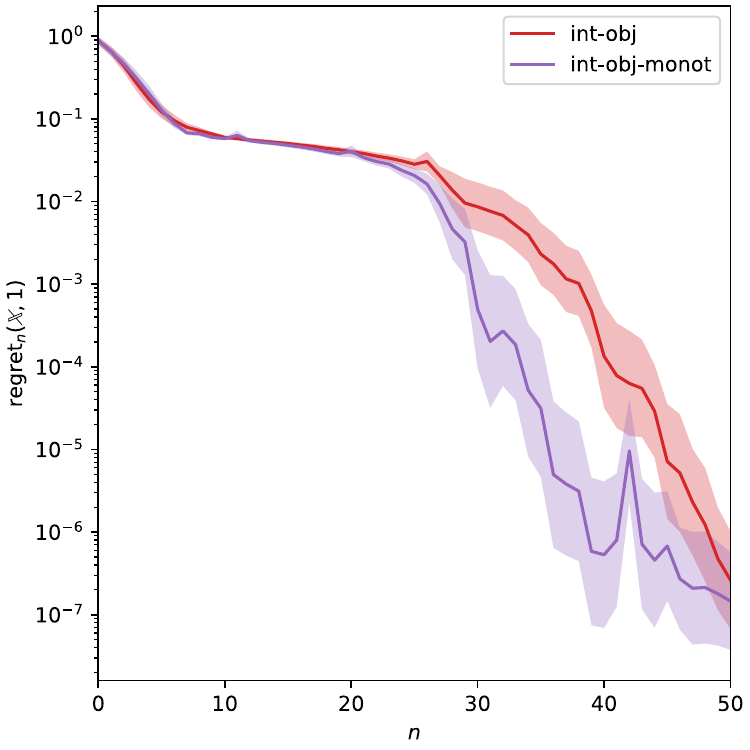}%
    \end{subfigure}
    \hfill{}
    \begin{subfigure}{0.32\textwidth}
    \caption{Car Cab Design (7,9)}
    \includegraphics[width=\textwidth]{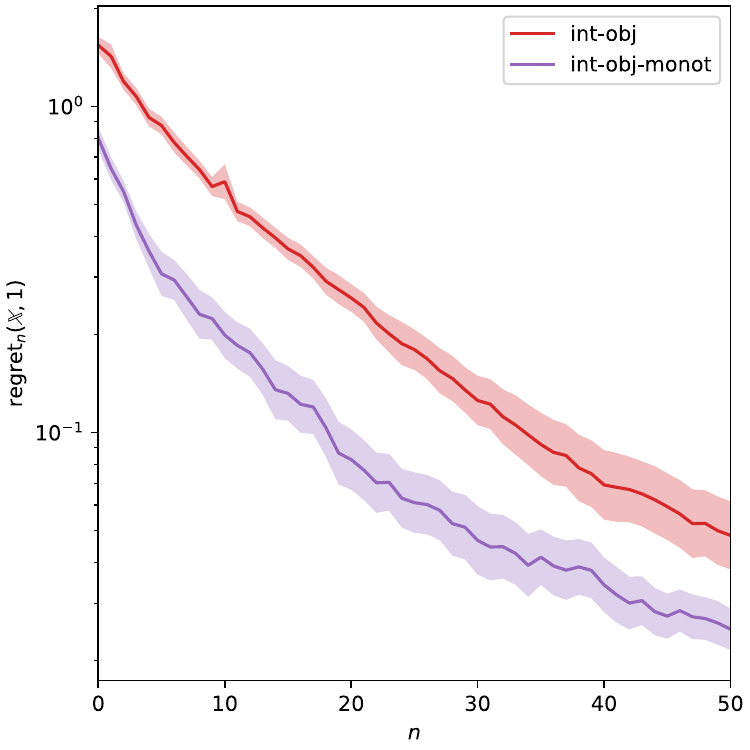}%
    \end{subfigure}
    \caption{Effect of the monotonicity for DTLZ2 with 9 decision variables and 6 objectives, WFG3 with 14 decision variables and 9 objectives, and Car Cab Design with 7 decision variables and 9 objectives.}
    \label{fig:mono}
\end{figure}

To enforce monotonicity, we use 64 pairs of points generated according to Algorithm \ref{alg:monot} with $\delta = 0.25$. The results in Figure~\ref{fig:mono} indicate that while the benefits of monotonicity are modest, they remain significant. This observation suggests that exploring more sophisticated approaches to induce monotonicity in the utility model could further improve performance, albeit with potentially higher computational costs. This improvement could be particularly valuable in high-dimensional objective spaces, where monotonicity may help avoid exploration of vast suboptimal regions far from the Pareto front.

\subsubsection{Noisy responses}
\label{sec:noise_results}
To assess the robustness of our method against noise in DM responses, we perform a reduced set of experiments where the simulated DM responses are corrupted by different levels of noise. The noisy responses are generated using a logistic likelihood (see Section~\ref{sec:model}), with a noise level parameter chosen such that the DM makes errors in approximately 15\% or 30\% of comparisons when comparing pairs of points from the top 1\% of utility function values within the optimization domain $\X$. These levels are estimated using a large number of random points within $\X$.

\begin{figure}
    \centering
    \captionsetup[sub]{margin={-15pt,0pt}} 
    \begin{subfigure}{0.32\textwidth}
    \caption{DTLZ7 (5,3)}
    \includegraphics[width=\textwidth]{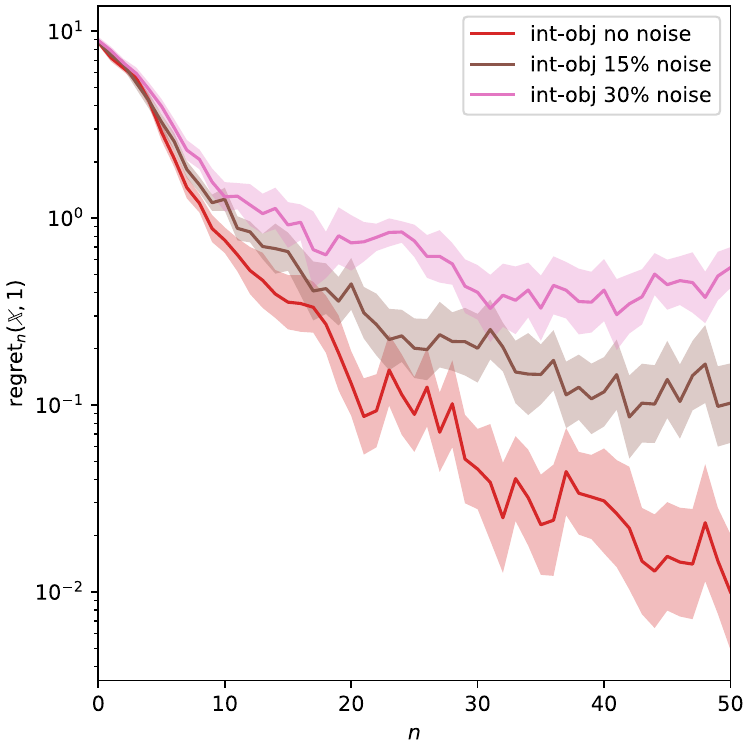}%
    \end{subfigure}
    \hfill{}
    \begin{subfigure}{0.32\textwidth}
    \caption{DTLZ2 (9,6)}
    \includegraphics[width=\textwidth]{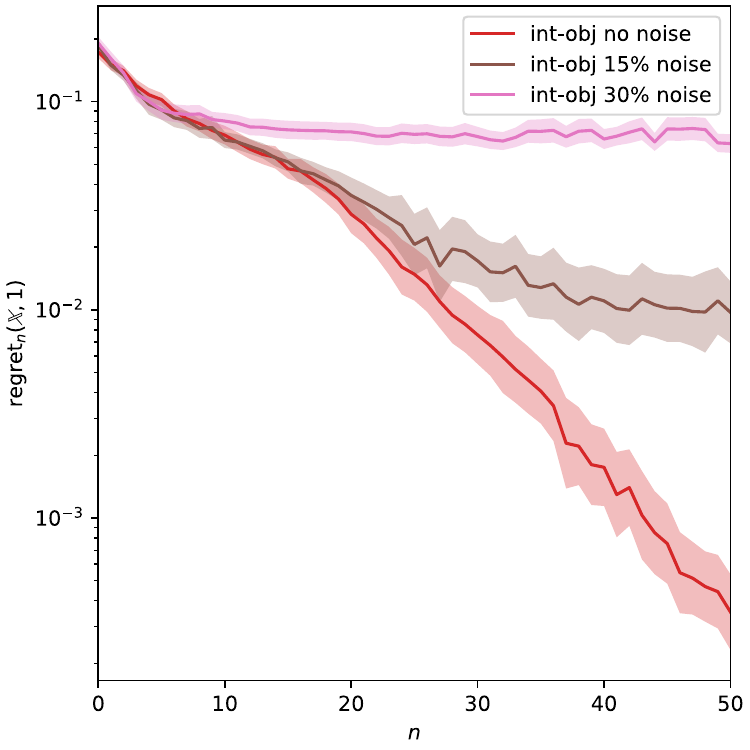}%
    \end{subfigure}
    \hfill{}
    \begin{subfigure}{0.32\textwidth}
    \caption{Car Cab Design (7,9)}
    \includegraphics[width=\textwidth]{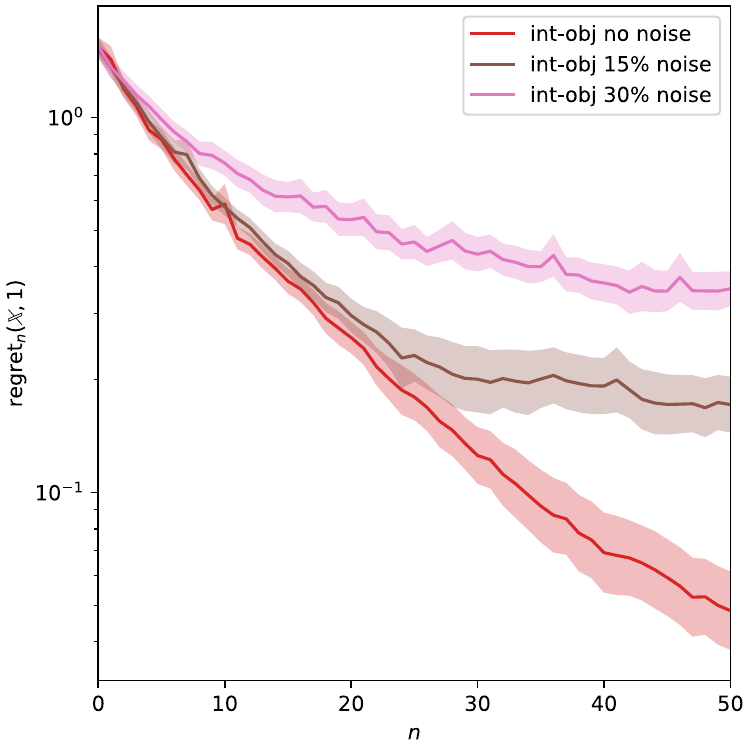}%
    \end{subfigure}
    \caption{Effect of noise for DTLZ7 with 5 decision variables and 3 objectives and DTLZ2 with 9 decision variables and 6 objectives, and Car Cab Design with 7 decision variables and 9 objectives. The noise levels corresponding to 0\%, 15\% and 30\% mistakes of the DM at the top 1\% of the utility values in the domain are indicated by the `no noise', `15\% noise', and `30\% noise' labels.    }
    \label{fig:noise}
\end{figure}
As shown in Figure~\ref{fig:noise}, while noise in the DM's responses slows the convergence rate,
the regret curves still exhibit a desirable trend, indicating our method's resilience. This pattern suggests that our probabilistic model compensates effectively for noisy interactions, maintaining a stable trajectory toward high-utility solutions.

\begin{figure*}[t]
    \centering
    \begin{subfigure}{0.32\textwidth}
    \caption{\makebox[0.8\textwidth][l]{DTLZ7 (5,3)}\hspace*{-100pt}}
    \includegraphics[width=\textwidth]{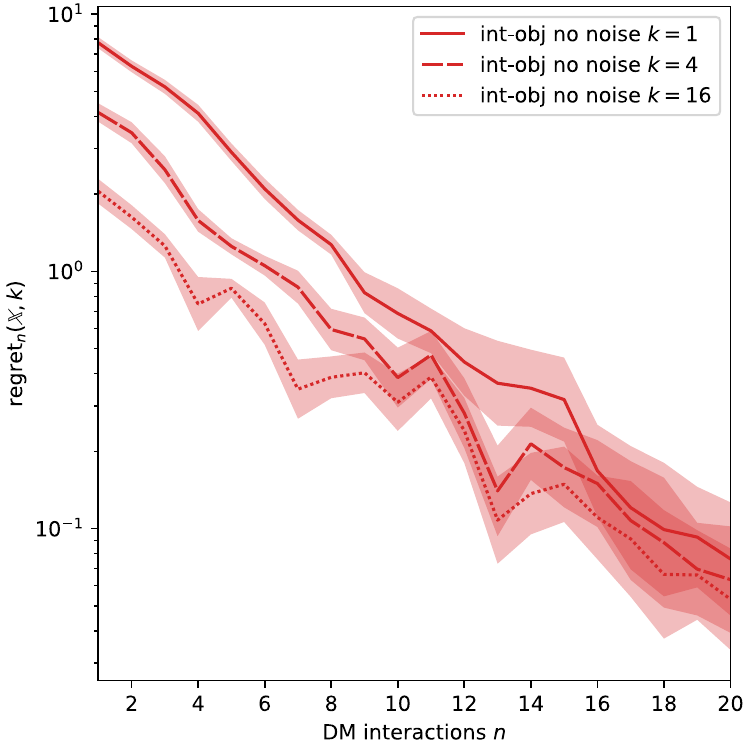}%
    \end{subfigure}
    \hfill{}%
    \begin{subfigure}{0.32\textwidth}
    \caption{\makebox[0.8\textwidth][l]{DTLZ2 (9,6)}\hspace*{-100pt}}
    \includegraphics[width=\textwidth]{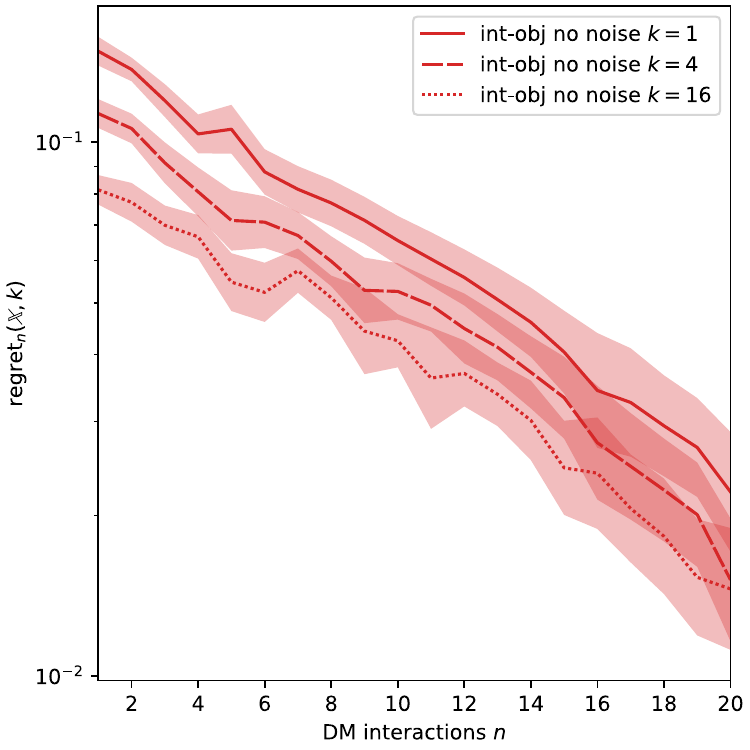}%
    \end{subfigure}
    \hfill{}%
    \begin{subfigure}{0.32\textwidth}
    \caption{\makebox[0.8\textwidth][l]{Car Cab Design (7,9)}\hspace*{-60pt}}
    \includegraphics[width=\textwidth]{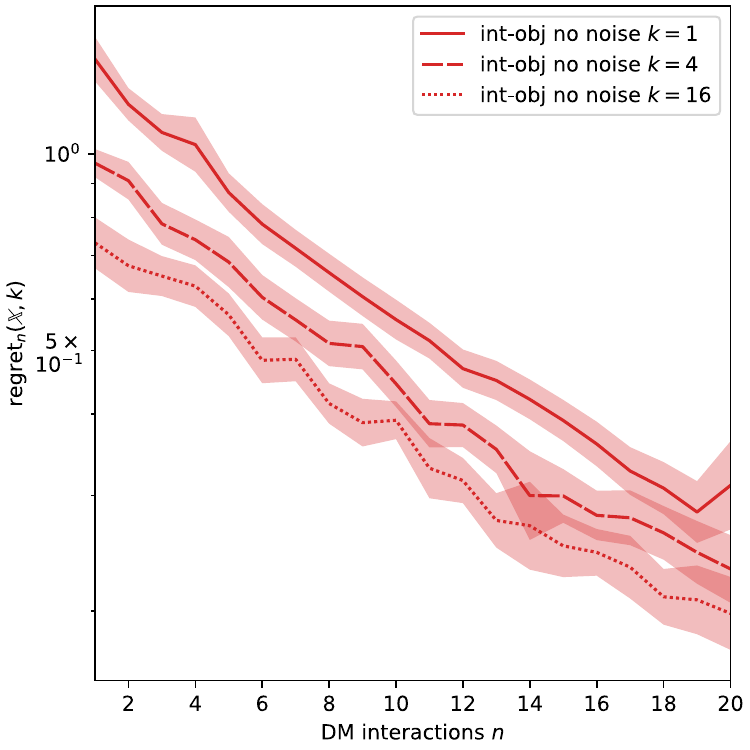}%
    \end{subfigure}
    \\[1.5ex]%
    \includegraphics[width=0.32\textwidth]{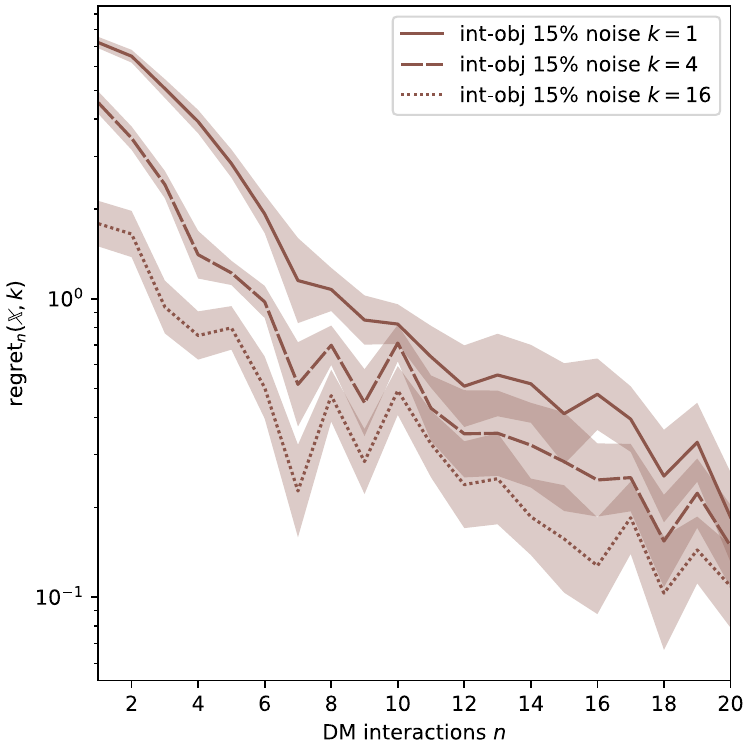}%
    \hfill{}%
    \includegraphics[width=0.32\textwidth]{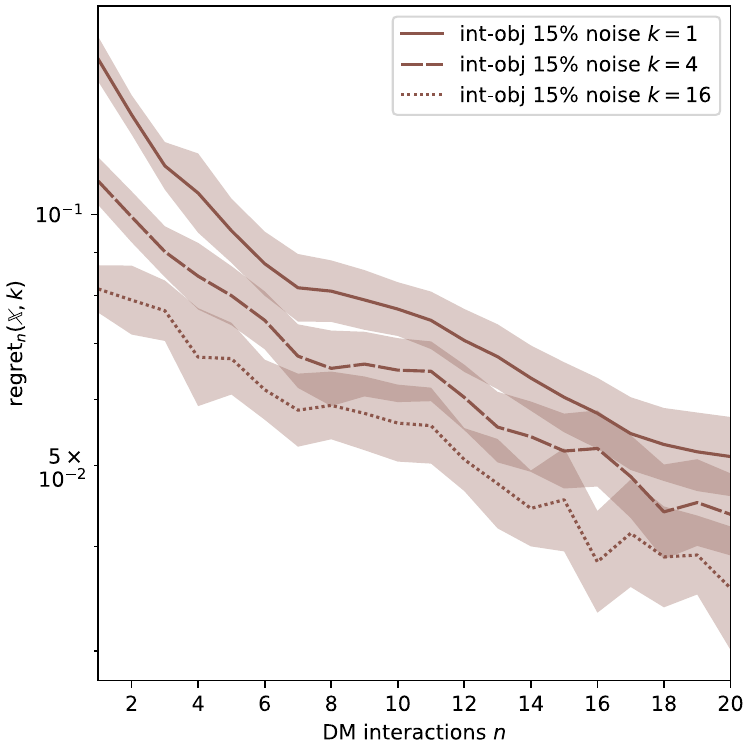}%
    \hfill{}%
    \includegraphics[width=0.32\textwidth]{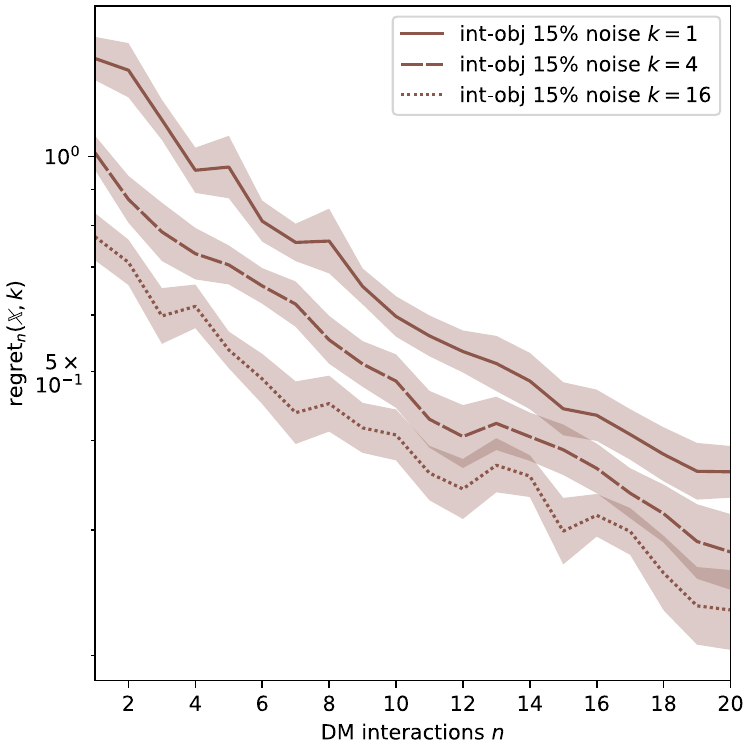}%
    \caption{Regret for different menu sizes for DTLZ7 with 5 decision variables and 3 objectives (left), DTLZ2 with 9 decision variables and 6 objectives (center), and Car Cab Design with 7 decision variables and 9 objectives (right). The first row shows results for noise-free responses from the DM. The second row shows results for noisy responses with medium noise level (i.e., 15\% mistakes of the DM at the top 1\% of the utility values in the domain). Short, medium, and small dashes correspond to menus of size 1, 4, and 16, respectively. }
    \label{fig:menu}
\end{figure*}

\subsubsection{Menu selection}

\label{sec:menu_results}
In this section, we evaluate the impact of different menu sizes, specifically $k = 1$, $4$, and $16$. To improve computational efficiency, we adopt a greedy construction strategy: Instead of jointly optimizing over all $k$ points, we build the menu incrementally by selecting one point at a time. At each step, we fix the previously selected points and optimize only the next one using Equation~\eqref{eq:menu}. This approach offers substantial computational savings while preserving strong theoretical guarantees; specifically, it achieves a $(1 - e^{-1})$-approximation to the optimal menu obtained via joint optimization under a standard submodularity argument \citep{Wilson2018Maximizing}.

As shown in Figure~\ref{fig:menu}, increasing the menu size generally leads to improved performance, especially when the number of DM interactions is limited or the responses are noisy. However, as $k$ grows, the same submodularity property implies diminishing returns, likely compounded by the increasing difficulty of the underlying optimization problem.

\section{Conclusion}
\label{sec:conclusion}
In this work, we proposed an approach to assist a DM in finding their most preferred solution within the Pareto set of an objective function. Our method assumes that the DM's preferences are represented by an underlying utility function, which we estimate from pairwise comparisons between solutions expressed by the DM. We employ a Bayesian framework using Gaussian processes to model the utility function and utilize an acquisition function to balance the exploration-exploitation trade-off when selecting queries for the DM. Furthermore, we demonstrated how to construct a menu of high-quality solutions after the elicitation phase, leveraging the uncertainty quantification from our probabilistic model. Through several experiments, we showed that our approach effectively identifies high-quality solutions according to the DM's preferences using only a small number of queries.

There are many exciting directions for future research. On the application side, our approach offers a principled way to find high-quality solutions in problems involving a larger number of objectives than those typically handled by traditional methods, potentially enabling new and impactful applications. On the methodological side, exploring alternative probabilistic models beyond GPs for estimating the DM's utility function would be of great interest. In particular, models that are both flexible and capable of incorporating prior structural information, such as monotonicity or concavity, could significantly enhance the efficiency and robustness of our method. Additionally, it would be valuable to investigate alternative frameworks for modeling DM preferences in scenarios where (i) assuming a utility function is not appropriate (e.g., when transitivity does not hold), and (ii) there are multiple stakeholders (e.g., when a single utility function may not suffice). Developing efficient elicitation strategies tailored to these alternative models could further expand the scope and applicability of preference-based multi-objective optimization.


\bmsection*{Acknowledgments}
Felix Huber was funded by Deutsche Forschungsgemeinschaft (DFG, German Research Foundation) under Germany's Excellence Strategy - EXC 2075 -- 390740016 and acknowledges the support by the Stuttgart Center for Simulation Science (SimTech).
Sebastian Rojas Gonzalez acknowledges support by FWO (Belgium) grant number 1216021N and the Belgian Flanders AI Research Program. 



\bibliography{bibl}


\newpage 
\appendix

\bmsection{Model details}

\vspace*{12pt}
\bmsubsection{Variational approximation of the posterior\label{app1.1a}}

For data consisting of preferences for pairwise comparisons the posterior distribution cannot be computed in closed form. Therefore, we approximate the posterior distribution $p(u' \given \D_n)$ using a variational approach with a Gaussian distribution $q(u') = \normaldist(u' \given \mu', \Sigma')$ at $\ninducing$ inducing points with corresponding utility values $u' \in \R^\ninducing$ as introduced in \cite{Hensman2015Scalable}. 

For this, we fit the mean $\mu' \in \R^\ninducing$ and covariance $\Sigma' \in \R^{\ninducing\times\ninducing}$ of the inducing points similarly to \cite{Nguyen2021TopK} by maximizing the evidence lower bound (ELBO)
\begin{equation*}
    \operatorname{ELBO}(q) = \int_{\IR^\ninducing} q(u') \log p(\D_n \given u') - q(u') \log\frac{q(u')}{p(u')} \dd u'
\end{equation*}
\noindent with the logistic likelihood (\ref{eq:likelihood}). The predictive distribution of the model is then a Gaussian process depending on $\mu'$ and $\Sigma'$ \citep{Nguyen2021TopK} and thus allows an efficient maximization of the acquisition function. 
Following \cite{Astudillo2023qEUBO}, we use all points from previous queries as inducing points and extend the set with quasi-random Sobol points to improve the model's numerical stability.
%




\end{document}